\documentclass[11pt]{article}
\usepackage{acl2015}
\usepackage{times}
\usepackage{url}
\usepackage{latexsym}
\usepackage{amssymb}
\usepackage{times}
\usepackage{url}
\usepackage{latexsym}
\usepackage{graphicx,epstopdf}
\usepackage{latexsym}
\usepackage[tight]{subfigure}
\usepackage{algorithm}
\usepackage{algorithmic}
\usepackage{amsmath} 
\usepackage{amssymb}  
\usepackage{cases}
\usepackage{collref}
\usepackage{float}
\usepackage{array}
\usepackage{xcolor}
\newcolumntype{C}[1]{>{\centering}p{#1}}

\title{Distant Supervision for Entity Linking}

\author{Miao Fan$^*$, Qiang Zhou and Thomas Fang Zheng\\
CSLT, Division of Technical Innovation and Development\\ Tsinghua National Laboratory for Information Science and Technology\\ Department of Computer Science and Technology, \\Tsinghua University, Beijing, 100084, China \\
\tt{$^*$fanmiao.cslt.thu@gmail.com}}

\date{}

\begin{document}
\maketitle
\begin{abstract}
Entity linking is an indispensable operation of populating knowledge repositories for information extraction. It studies on aligning a textual entity mention to its corresponding disambiguated entry in a knowledge repository. In this paper, we propose a new paradigm named distantly supervised entity linking (DSEL), in the sense that the disambiguated entities that belong to a huge knowledge repository (Freebase) are automatically aligned to the corresponding descriptive webpages (Wiki pages). In this way, a large scale of weakly labeled data can be generated without manual annotation and fed to a classifier for linking more newly discovered entities. Compared with traditional paradigms based on solo knowledge base, DSEL benefits more via jointly leveraging the respective advantages of Freebase and Wikipedia. Specifically, the proposed paradigm facilitates bridging the disambiguated labels (Freebase) of entities and their textual descriptions (Wikipedia) for Web-scale entities. Experiments conducted on a dataset of 140,000 items and 60,000 features achieve a baseline F1-measure of 0.517. Furthermore, we analyze the feature performance and improve the F1-measure to 0.545.
\end{abstract}
\section{Introduction}
To build the ``Digital Alexandria Library'' for our human race, researchers in the NLP community have dedicated themselves to {\it Information Extraction} \cite{sarawagi2008information} over the past decades. Information extraction focuses on processing natural language text to produce structured knowledge, which is usually represented as triples (two entities and their relation) for the convenience of storage in a database, retrieval, or even automatic reasoning. For example, if we send a natural language sentence, {\it Michael Jordan visited CMU yesterday}, to the pipeline of information extraction machine, it will be processed by three operations in advance, i.e.,
\begin{itemize}
  \item {\bf Named Entity Recognition} \cite{Nadeau2007}: Entities should firstly be identified and classified into predefined categories, such as person (PER), location (LOC) and organization (ORG). The sentence will be annotated as {\it [Michael Jordan]/PER visited [CMU]/ORG yesterday}, after being processed by this operation.
  \item {\bf Coreference Resolution} \cite{Ng2010}: Some entities may have alias or abbreviations. It is well known that {\it CMU} is the abbreviation for {\it Carnegie Mellon University}. The knowledge repository may only store the regularized name, e.g., {\it Carnegie Mellon University}, for this named entity, so coreference resolution is indeed necessary.
  \item {\bf Relation Extraction} \cite{Bach2007}: After both of the named entities ({\it [Michael Jordan]/PER} and {\it [Carnegie Mellon University]/ORG}) are recognized and regularized, we begin to study on the relation between them. In this case, we extract the verb {\it visited} and map it to the relation {\it visit}. Then the output will be a triple, i.e., ({\it Michael Jordan [PER], visit, Carnegie Mellon University [ORG]}).
\end{itemize}

So far, we only abstract the triple as the structured knowledge from the natural language sentence. However, it devotes nothing to increasing the scale of the knowledge repository such as Freebase \cite{Bollacker2007}  which is a huge\footnote{According to the statistics released on 10th March, 2014 by Google Inc., there are about 1.9 billion Freebase triples and 43 million entities.}, public\footnote{The whole dump of Freebase can be downloaded from \url{https://developers.google.com/freebase/data}}, collaborative\footnote{One can access to Freebase and contribute more knowledge.}\cite{Bollacker2008} and online knowledge base with billions of triples and millions of disambiguated entities, and is primarily maintained by Google Inc., because we even do not know which exact {\it Michael Jordan} the triple ({\it Michael Jordan [PER], visit, Carnegie Mellon University [ORG]}) refers to in Freebase. As illustrated in Figure 1, there are three different persons named {\it Michael Jordan} in Freebase and each of them may be the protagonist of that news. Therefore, to populate knowledge repositories \cite{Ji2011}, we need the {\it fourth operation}:
\begin{itemize}
  \item {\bf Entity Linking} \cite{Rao2013}: It concerns about the study of aligning a textual entity mention to the corresponding disambiguated entry in a knowledge repository. More specifically, since there are several {\it Michael Jordan} disambiguated by different MIDs (machine identifiers) as illustrated in Figure 1, we may build a classifier that can help assign the {\it Michael Jordan} in the extracted triplet {\it (Michael Jordan [PER], visit, Carnegie Mellon University [ORG])} to the exact named entity in Freebase or find out that this {\it Michael Jordan} is a newly discovered named entity (NIL).
\end{itemize}

\begin{figure}
\centering
\includegraphics[width=0.5\textwidth]{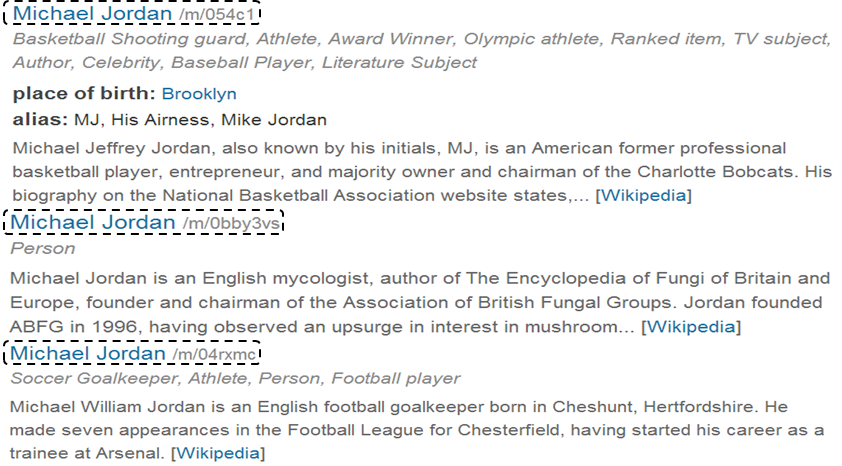}\\
\caption{The disambiguated entities with the same name {\it Michael Jordan} in Freebase. The entities in Freebase are disambiguated by a unique machine identifier, e.g., the famous basketball player, Michael Jordan labeled by 054c1 (MID).}

\end{figure}
Hachey et al. \shortcite{Hachey2013} and Rao et al. \shortcite{Rao2013} elucidate that most of the literatures \cite{Bunescu2006,Mihalcea2007,Cucerzan2007,Milne2008,Ratinov2011} and the entity linking tracks\footnote{\url{http://www.nist.gov/tac/2013/KBP/EntityLinking/index.html}} in TAC-KBP \cite{McNamee2009,Ji2010} concentrate on linking ambiguous entities to the entries in Wikipedia, whereas our ultimate goal is to populate the structured knowledge repository, e.g., Freebase. However, to the best of our knowledge, few works \cite{Zheng2012} concern about disambiguating named entities using Freebase which contains much more entries but less text information for each entry than Wikipedia.

Overall, Hachey et al. \shortcite{Hachey2013} and Zheng et al. \shortcite{Zheng2012} represent two research directions leveraging Wikipedia and Freebase, respectively. As both of the two collaborative web resources have their respective superiorities, i.e., more context information and more disambiguated entities, we begin to study a new paradigm that could bridge the gap between those two separated repositories and benefit from their respective advantages. From the perspective of supervised learning, entity linking can be naturally regarded as a classification problem. To build a training dataset for disambiguating a set of entities with the same name, we can firstly collect the sentences that mention that name from webpages, such as Wiki pages\footnote{The Wiki page for the famous basketball player, Michael Jordan, is \url{http://en.wikipedia.org/wiki/Michael_jordan}.}, and then manually annotate each entity mention with its unique machine identifier (MID) in Freebase given the contexts of sentences that it occurs in. However, hand-labeled data is time consuming and usually applicable to some specific classes of entities, such as person ({\it PER}), location ({\it LOC}) and organization ({\it ORG}).
Therefore, we look forward to an approach that averts the tedious and laborious work.
\begin{figure}
\centering
\includegraphics[width=0.5\textwidth]{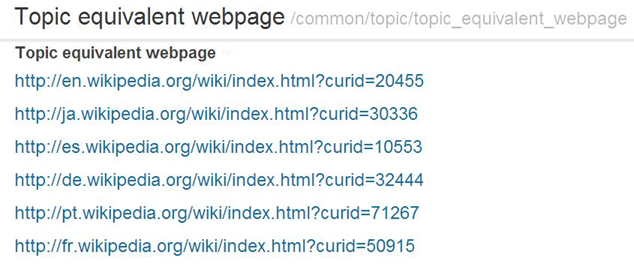}\\
\caption{The topic equivalent webpages of the famous basketball player, {\it Michael Jordan} in Freebase.}
\end{figure}

Inspired by the idea of weak labeling \cite{Fan2014,Craven1999}, we contribute a new paradigm called distantly supervised entity linking (DSEL) without manual annotation in this paper. More specifically, we take advantage of a heuristic alignment assumption based on crowd sourcing to connect a certain disambiguated entity in Freebase with its related webpages. In these webpages, feature vectors can be extracted from the sentence-level textual contexts of that entity mention, and be labeled by its corresponding MID in Freebase. Then we can produce a large scale of weakly labeled\footnote{Auto-labeling via crowd sourcing may naturally bring about noise. Therefore, we regard the dataset weakly labeled.} dataset in this way. Moreover, it is unrealistic to learn a specific classifier for each entity, as there are about 43 million disambiguated entities in Freebase. To tackle with those challenges, we propose a strategy of training a general classifier for disambiguating multiple entities and select a well known classifier, i.e., {\it liblinear} \cite{Fan2008} to self-learn the weights among the high-dimensional sparse and noisy features.
Experiments are conducted on a dataset of 140,000 items and 60,000 features. DSEL achieves a baseline F1-measure of 0.517. Furthermore, we analyze the performance influenced by other different features, and finally the F1-measure is improved to 0.545.

\begin{figure*}
  \centering
  \includegraphics[width=1.0\textwidth]{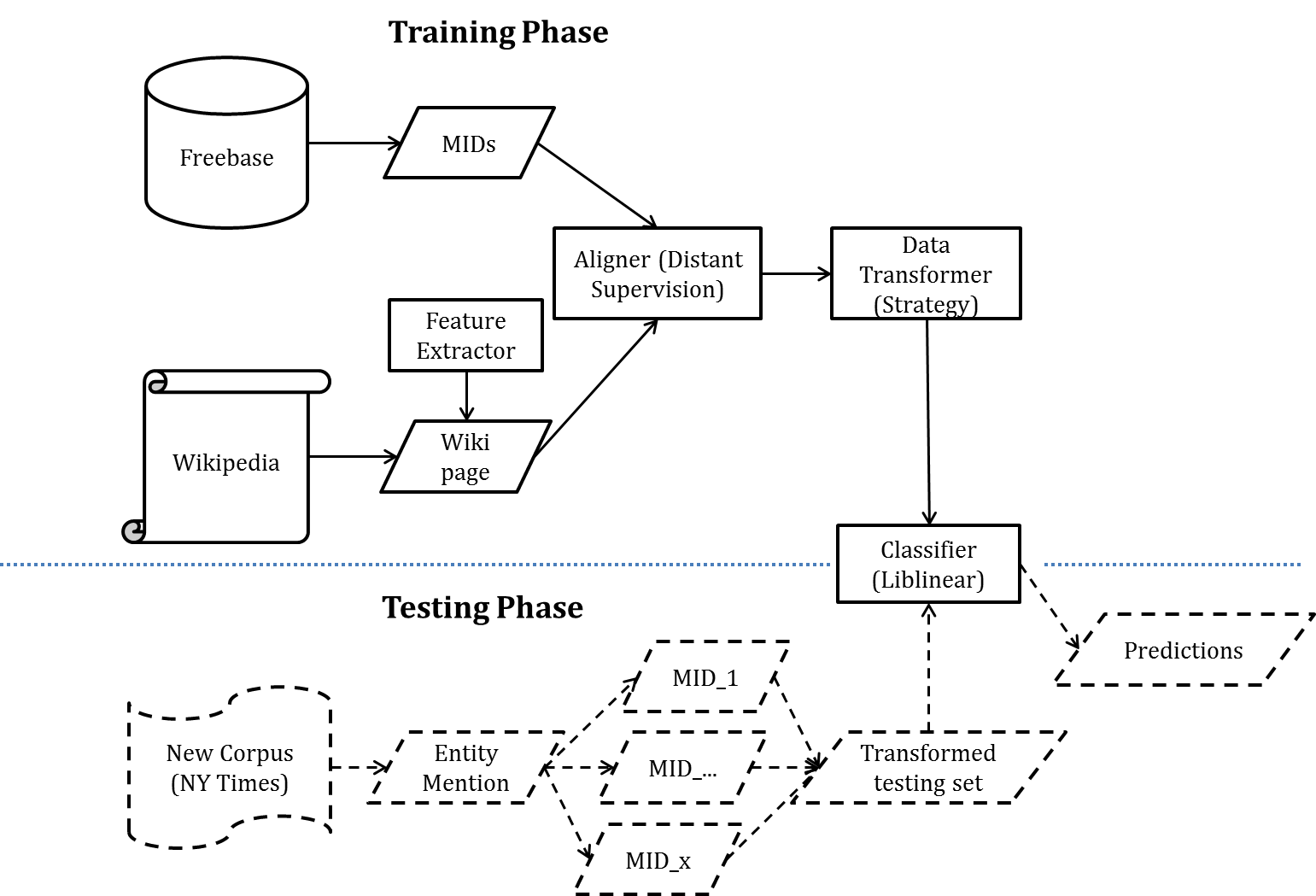}\\
  \caption{The architecture of DSEL system.}

\end{figure*}

\begin{table*}[!htp]
\centering
\begin{tabular}{|C{3.5cm}|C{11cm}|}
\hline
{\bf Sentence} & {\it His biography on the National Basketball Association (NBA) website states,  ``By acclamation, {\bf Michael Jordan} is the greatest basketball player of all time.''}\tabularnewline
\hline
{\bf BOW ($K = 1$)} &  $<$\{acclamation\}, \{is\}$>$ \tabularnewline
{\bf BOW ($K = 2$)} &  $<$\{states\}, \{acclamation\}, \{is\}, \{greatest\}$>$ \tabularnewline
{\bf BOW ($K = 3$)} &  $<$\{website\}, \{states\}, \{acclamation\}, \{is\}, \{greatest\}, \{basketball\}$>$ \tabularnewline

\hline
{\bf WS ($K = 1$)} &  $<$\{acclamation\}, \{is\}$>$ \tabularnewline
{\bf WS ($K = 2$)} &  $<$\{states-acclamation\}, \{is-greatest\}$>$ \tabularnewline
{\bf WS ($K = 3$)} &  $<$\{website-states-acclamation\},\{is-greatest-basketball\}$>$ \tabularnewline

\hline
{\bf BOW + POS ($K = 1$)} &  $<$\{acclamation/NN\}, \{is/VBZ\}$>$ \tabularnewline
{\bf BOW + POS ($K = 2$)} &  $<$\{states/NNS\}, \{acclamation/NN\}, \{is/VBZ\} \{greatest/JJS\}$>$ \tabularnewline
{\bf BOW + POS ($K = 3$)} &  $<$\{website/NN\}, \{states/NNS\}, \{acclamation/NN\}, \{is/VBZ\}, \{greatest/JJS\}, \{basketball/NN\}$>$ \tabularnewline

\hline
{\bf WS + POS ($K = 1$)} &  $<$\{acclamation/NN\}, \{is/VBZ\}$>$ \tabularnewline
{\bf WS + POS ($K = 2$)} &  $<$\{states/NNS-acclamation/NN\}, \{is/VBZ-greatest/JJS\}$>$ \tabularnewline
{\bf WS + POS ($K = 3$)} &  $<$\{website/NN-states/NNS-acclamation/NN\},\{is/VBZ-greatest/JJS-basketball/NN\}$>$ \tabularnewline

\hline
\end{tabular}
\caption{Twelve kinds of lexical features for the given sentence. A pair of angle brackets stands for a feature vector, e.g., {\it $<$\{states\}, \{acclamation\}, \{is\} \{greatest\}$>$}. A feature item is marked by a pair of braces, e.g.,  {\it \{states-acclamation\}}.}
\end{table*}
\section{Paradigm}

Traditional supervised learning methods for entity disambiguation require tedious labor on manual annotation to build training datasets. Manual annotation costs a lot, and can only cover some specific category, e.g., person names \cite{Christen2006} as well. Therefore, we look forward to exploring a paradigm that could automatically generate large scale of open-category training datasets without manual annotation. Based on the dataset, we aim to build a practical classifier and generalize it to disambiguate more unlinked entity mentions in free texts.

Freebase contains 43 million disambiguated entities falling into 76 categories. Each entity is assigned by a unique machine identifer (MID). Those MIDs are the natural labels for the newly identified entity mentions linking to. However, there are inadequate free texts {\it locally} for extracting features, as Freebase is a well-structured knowledge repository with billions of triples. Therefore, we resort to other free-text corpus that could be {\it distantly supervised} by Freebase and the key challenge is to find the bridge of supervision.

Fortunately, every entity in Freebase maintains a list of links to its topic equivalent webpages via crowd sourcing \cite{Howe2006} as shown in Figure 2. These links will guide us to find the description webpages for that entity. Even though those links involves in different languages, we only choose the English Wiki pages to conduct experiments.
Overall, we jointly exploit Freebase and Wikipedia to automatically construct the data for training a classifier.

\section{Feature}
For each entity in Freebase, we find its topic-equivalent Wiki page and extract the contextual features of its mention at sentence level.

Generally, we simultaneously choose $ K$ ($K = 1, 2, 3$) open-class words \cite{VanPetten1991}, namely nouns, verbs, adjectives and adverbs, in front and behind the given entity mention. If we ignore the sequence of these words, we can gain the bag-of-words feature, whereas the word sequence feature. Furthermore, we use Stanford NLP core\footnote{\url{http://nlp.stanford.edu/software/corenlp.shtml}} and add the part-of-speech tagging feature which may help disambiguate those contextual words. Therefore, for each $ K$ size window surrounding the entity mention, we could extract four kinds of different features, i.e., bag of words (BOW), word sequence (WS), bag of words plus part-of-speech tagging (BOW + POS) and word sequence plus part-of-speech tagging (WS + POS). In total, there are twelve kinds of lexical features.

To elucidate the various kinds of contextual features, we randomly pick up a sentence from the Wiki page of the famous basketball player as example, i.e.,

{\it His biography on the National Basketball Association (NBA) website states,  ``By acclamation, {\bf Michael Jordan} is the greatest basketball player of all time.''}

The twelve kinds of lexical features for the sentence above are listed in Table 1. We will compare the performance among these features in Section 5.

\section{Implementation}
As we have already automatically produced a training dataset based on the proposed distant supervision paradigm, an intuitive idea is to feed a specific classifier for each ambiguous name with its unambiguous MIDs and the corresponding feature vectors. However, Table 2 shows that there are at least 5.5 million names that denominate more than one entity (MID) in Freebase. Therefore, it is infeasible to build 5.5 million specific classifiers.
\begin{table*}
\centering
\begin{tabular}{|C{3cm}|C{1.5cm}|C{3cm}|C{1.5cm}|C{3cm}|C{1.5cm}|}
\hline
\# of MIDs with the same name & \# of names & \# of MIDs with the same name & \# of names & \# of MIDs with the same name & \# of names
\tabularnewline
\hline
\hline
{\bf 2} & 4,467,216 & {\bf 5} & 180,489 & {\bf 8} & 60,273
\tabularnewline
{\bf 3} & 740,530 & {\bf 6} & 134,012 & {\bf 9} & 41,256
\tabularnewline
{\bf 4} & 440,261 & {\bf 7} & 76,459 & {\bf 10} & 33,628
\tabularnewline
\hline
\end{tabular}
\caption{The distribution of ambiguous entities in Freebase.}
\end{table*}
To train a general classifier that does not restrict itself to disambiguating a certain name, we adopt a strategy that merges those specific classifiers. Concretely, we transform MIDs, the original labels into features and use 1/0 to indicate whether the contextual features from Wiki pages and MIDs in Freebase match or not with each other. If we choose the BOW ($K = 3$) feature in Table 1 for instance, one positive training sample will contain a new feature vector ($<$\{website\}, \{states\}, \{acclamation\}, \{is\}, \{greatest\}, \{basketball\}, \{MID:054c1\}$>$) labeled by 1. To balance the training dataset, we randomly pick up features from other entities uniformly named to generate negative samples. For example, another well-known {\it Michael Jordan} (MID:0bby3vs) is an English mycologist. We can extract a BOW ($K = 3$) feature vector, i.e., $<$\{is\}, \{English\}, \{mycologist\}$>$ , and it concatenates \{MID:054c1\} to construct a negative sample labeled by 0.

The distant supervision paradigm and the strategy of building the training set for a general classifier lead to high-dimensional noisy and sparse features. Moreover, given the millions of training samples produced by aligning Freebase and Wikipedia, we choose a linear classifier that is based on logistic regression approach, i.e., Liblinear \cite{Fan2008}, to rapidly self-learn the weights among the high-dimensional sparse and noisy features.

For a newly discovered entity mention in the testing corpus, we firstly extract its contextual feature, e.g., bag of words as above. Then the feature concatenates all the candidate MIDs that share the same name with that entity mention. Each testing sample within the same name collection will predict a score indicating the strength of linking. For each collection, the Top-N predictions with higher probabilities are selected for evaluation.

We summarize the procedures of implementing our proposed paradigm and use Figure 3 to demonstrate the architecture of DSEL system.

\section{Experiments}
In this section ,we report the experimental results following the procedures described in Section 4. To evaluate the performance of different features, we adopt three widely used metrics \cite{Meij2013}, namely precision, recall and F1-measure.

\subsection{Dataset}
We randomly select 20,000 ambiguous names (collections) in Freebase. About 82,000 sentences that contain at least one entity mention are extracted from the topic-equivalent Wiki pages. For each collection, 80\% sentences are randomly picked up for constructing the training set and 20\% remains are for held-out evaluation. Following the procedures of building training samples described in Section 4, we gain a dataset including around 140,000 items and 60,000 features.

\subsection{Evaluation metrics}
Precision and recall are widely used metrics to evaluate different rank-based approaches on entity linking. F1-measure synthetically measures precision and recall by calculating the harmonic mean of them. Suppose that $C$ denotes the whole collection set for testing. $C_{i,j}$ represents the set of Top-$j$ predictions with higher probabilities in the $i$-th collection. $G_{i}$ stands for the set of gold standards of the $i$-th collection. $\#(S)$ is the function that counts the entries in set $S$. Then the formulae to calculate precision, recall and F1-measure are as follows,
\begin{table*}
\centering
\begin{tabular}{|C{3.5cm}|C{2.7cm}|C{3.5cm}|C{2.7cm}|}
\hline
 Feature type & Avg. F1-measure &  Feature type & Avg. F1-measure
\tabularnewline
\hline
\hline
{\bf BOW ($K = 1$)} & {\bf 0.539} & {\bf WS ($K = 1$)} & {\bf 0.544}
\tabularnewline
{\bf BOW ($K = 2$)} & 0.531 & {\bf WS ($K = 2$)} & 0.532
\tabularnewline
{\bf BOW ($K = 3$)} & 0.529 & {\bf WS ($K = 3$)} & 0.518
\tabularnewline
\hline
{\bf BOW + POS ($K = 1$)} & {\bf 0.540} & {\bf WS + POS ($K = 1$)} & {\bf 0.545}
\tabularnewline
{\bf BOW + POS ($K = 2$)} & 0.532 & {\bf WS + POS ($K = 2$)} & 0.531
\tabularnewline
{\bf BOW + POS ($K = 3$)} & 0.529 & {\bf WS + POS ($K = 3$)} & 0.517
\tabularnewline
\hline
\end{tabular}
\caption{The F1-measure comparison among different features.}
\end{table*}

\begin{figure}
  \center
  \subfigure[Precision-Recall curves for BOW features.]{
    \raisebox{-1cm}
    {\includegraphics[width=0.5\textwidth]{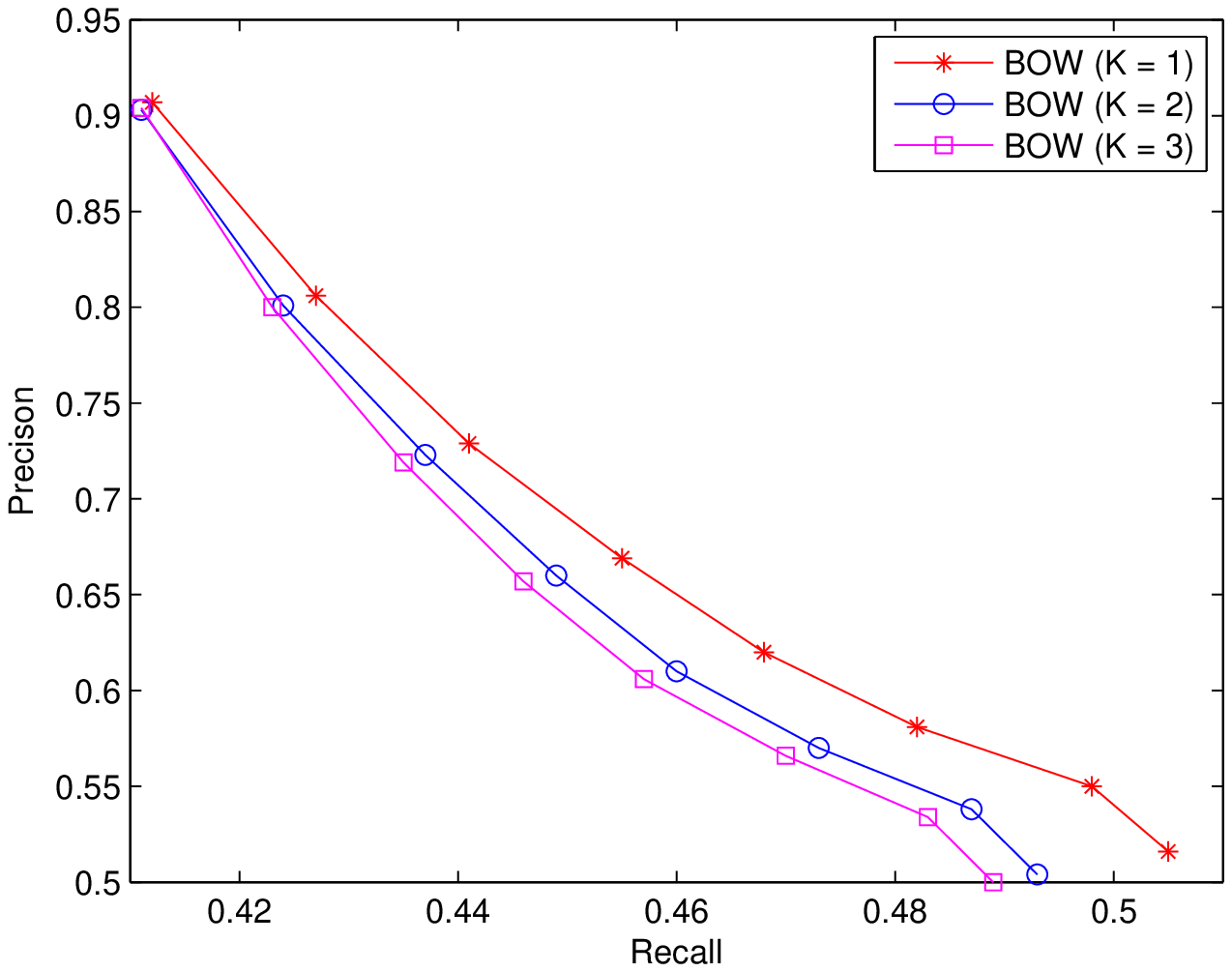}}
    }

      \subfigure[Precision-Recall curves for BOW + POS features.]{
    \raisebox{-1cm}{
    \includegraphics[width=0.5\textwidth]{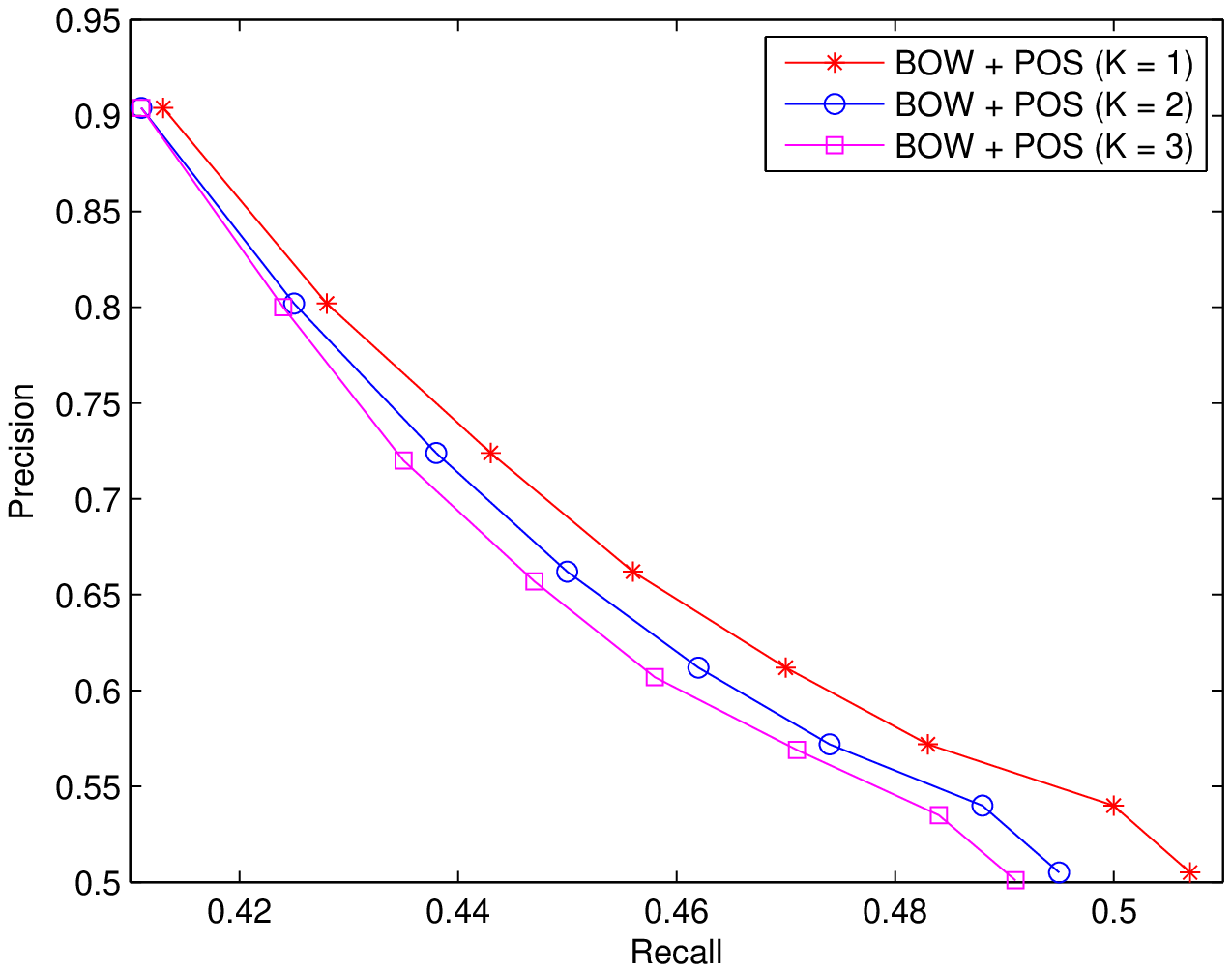}}
    }
\vspace{-5mm}
  \caption{Precision-Recall curves for the BOW-class lexical features.}

\end{figure}

\begin{figure}
  \center
    \subfigure[Precision-Recall curves for WS features.]{
        \raisebox{-1cm}{
    \includegraphics[width=0.5\textwidth]{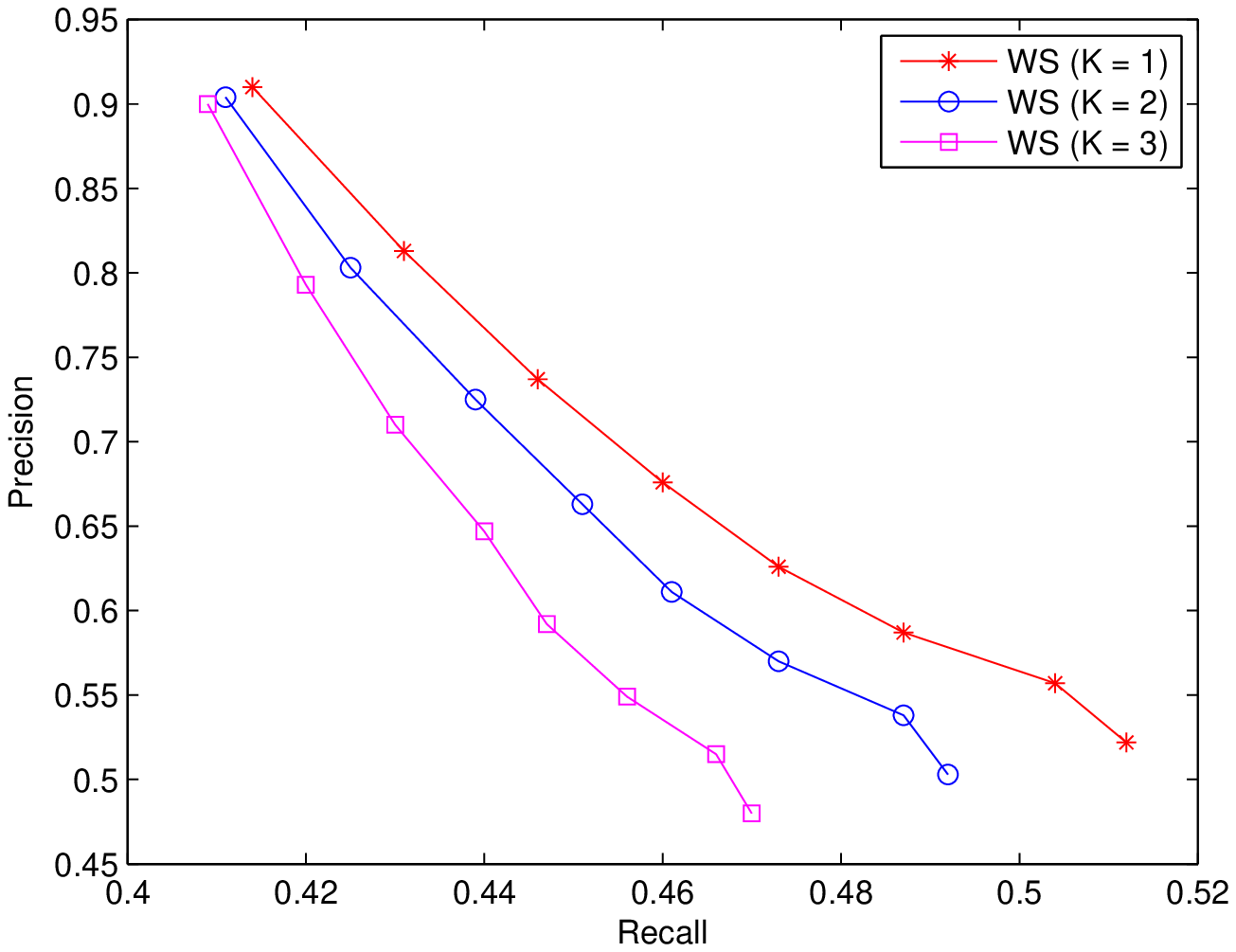}}

    }
    \subfigure[Precision-Recall curves for WS + POS features.]{
        \raisebox{-1cm}{
    \includegraphics[width=0.5\textwidth]{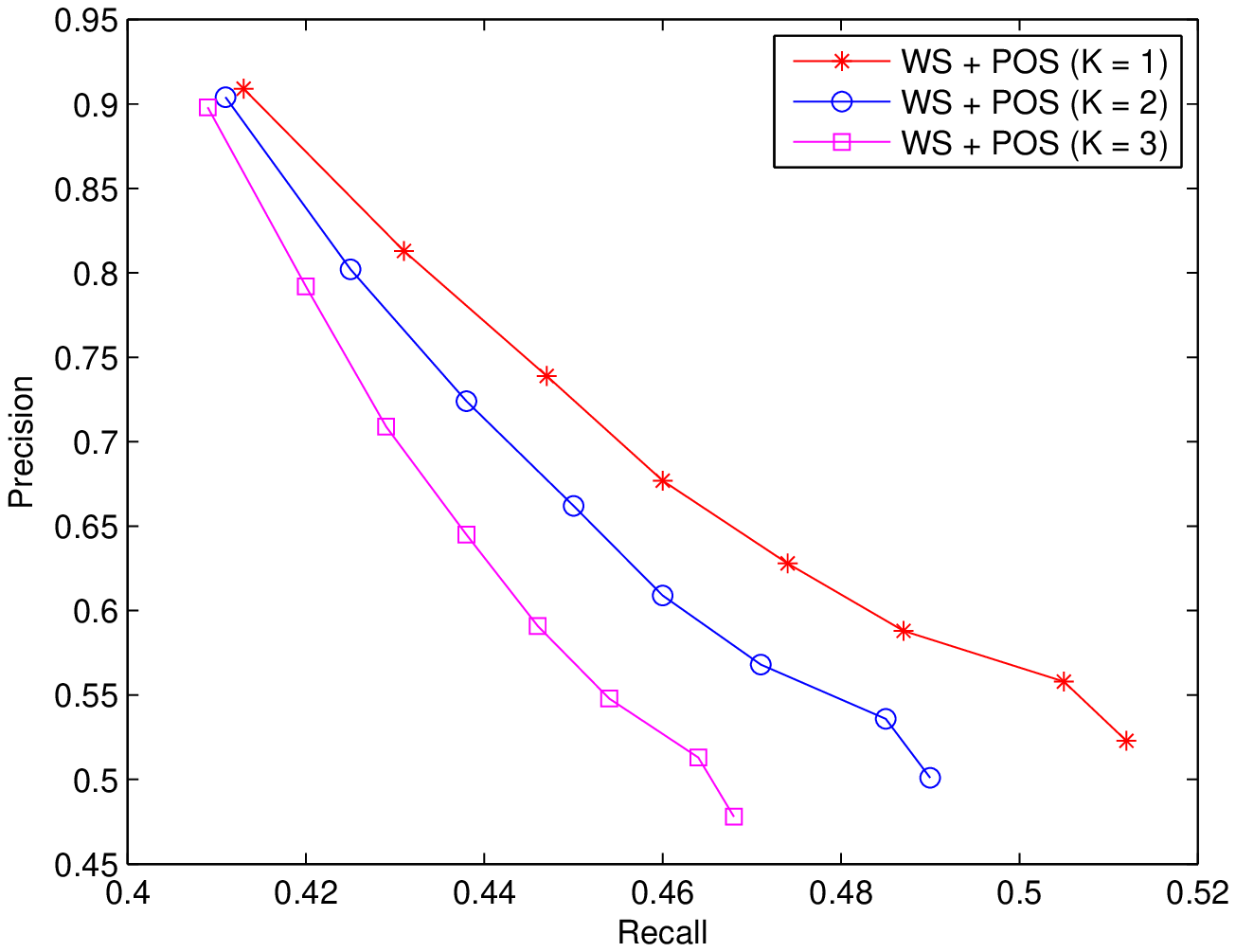}
    }}
\vspace{-5mm}
  \caption{Precision-Recall curves for the WS-class lexical features.}

\end{figure}

$$
\text {Precision} = \sum_{i}\sum_{j}\frac{{\# (C_{i, j} \bigwedge \ G_i)}} {\#(C_{i, j})},
$$

$$
\text {Recall} = \sum_{i}\sum_{j}\frac{{\# (C_{i, j} \bigwedge \ G_i)}} {\#(C)},
$$

$$
\text {F1-measure} = \frac{2 \times Precision \times Recall}{Precision + Recall}.
$$

\subsection{Feature comparison}
For each type of feature, we conduct one trial and tune the parameters for the logistic classifier using 5-fold cross validation. Then we adopt held-out testing taking advantage of the 20\% sentences left.

Figure 4 and Figure 5 show the precision-recall curves for the twelve lexical features, and Table 3 displays the average F1-measure comparison among different features. We find out that the WS-class features generally outperform the BOW-class features, and the short-distance contextual features ($K = 1$) are more effective than the long-distance ones ($K = 2, 3$).

\section{Conclusion and Future Work}
As far as we know, it is the first attempt to deal with the task of entity linking based on the idea of distant supervision. We leverage a heuristic alignment assumption, i.e., the topic equivalent pages, to bridge the gap between Freebase and Wikipedia and jointly use those two knowledge bases to automatically produce training data without manual annotation. Moreover, we propose a strategy that transforms labels into features and feed them to a general classifier, rather than building an individualized classifier for each ambiguous name for millions of entities.

For the future work, we believe that this new paradigm leaves several open questions:
\begin{itemize}
  \item Besides the entities (MIDs) that have already been stored in knowledge repositories (Freebase), new entity instances (NIL) with the same name need to be discovered. Therefore, further study could focus on extending paradigm to identify unknown entities.
  \item The link for many other webpages in different languages are also provided in Freebase, as illustrated in Figure 2. It may facilitate the research of cross-lingual entity linking.
  \item The alignment assumption is simple and heuristic. Further studies may dedicate on discovering other reasonable alignment principles.
  \item Even though the strategy for generating training data that fits a general classifier, it rises the problem that high-dimensional sparse and noisy features impact the effectiveness and efficiency of the proposed paradigm.
\end{itemize}

Generally speaking, the experiments prove that our new proposed paradigm is promising and it is worthy of being further studied.
\section*{Acknowledgements}
This work is mainly supported by National Program on Key Basic Research Project (973 Program) under Grant 2013CB329304, National Science Foundation of China (NSFC) under Grant No. 61373075. Thanks to Yulong Gu, Yingnan Xiao and anonymous reviewers for their insightful comments.

\bibliographystyle{acl}
\bibliography{acl2015}

\end{document}